\pdfoutput=1

\documentclass[11pt]{article}

\usepackage[]{EMNLP2023}

\usepackage{times}
\usepackage{latexsym}

\usepackage[T1]{fontenc}

\usepackage[utf8]{inputenc}

\usepackage{microtype}

\usepackage{graphicx}
\usepackage{indentfirst}
\usepackage{multirow}
\usepackage{multicol}
\usepackage{booktabs}
\usepackage{amsmath}
\usepackage{adjustbox}

\title{Why Antiwork: A RoBERTa-Based System for Work-Related Stress Identification and Leading Factor Analysis}

\author{Tao Lu, Muzhe Wu, Xinyi Lu, Siyuan Xu, Shuyu Zhan, Anuj Tambwekar, \\ {\bf and Emily Mower Provost} \\
  Department of Computer Science and Engineering,\\
  University of Michigan, Ann Arbor, MI, USA\\
  \texttt{\{luttul,henrw,lwlxy,xsyarctg,zsynexus,anujt,emilykmp\}@umich.edu}
}

\begin{document}
\maketitle
\begin{abstract}


Harsh working environments and work-related stress have been known to contribute to mental health problems such as anxiety, depression, and suicidal ideation. As such, it is paramount to create solutions that can both detect employee unhappiness and find the root cause of the problem.  While prior works have examined causes of mental health using machine learning, they typically focus on general mental health analysis, with few of them focusing on explainable solutions or looking at the workplace-specific setting. r/antiwork is a subreddit for the antiwork movement, which is the desire to stop working altogether. Using this subreddit as a proxy for work environment dissatisfaction, we create a new dataset for antiwork sentiment detection and subsequently train a model that highlights the words with antiwork sentiments. Following this, we performed a qualitative and quantitative analysis to uncover some of the key insights into the mindset of individuals who identify with the antiwork movement and how their working environments influenced them. We find that working environments that do not give employees authority or responsibility, frustrating recruiting experiences, and unfair compensation, are some of the leading causes of the antiwork sentiment, resulting in a lack of self-confidence and motivation among their employees.

\end{abstract}

\section{Introduction}
Toxic workplaces have been an important source of mental health (MH) problems \cite{alsomaidaee2023toxic}. Due to reasons such as unpleasant working environments, unreasonably designed workloads, and harsh supervisors, employees tend to develop a resistive feeling towards work, i.e., the \textbf{antiwork sentiment}. The antiwork sentiment can cause harmful negative emotions like dissatisfaction, frustration, and even irritation, which can put employees at risk of MH problems such as depression, burnout, sleep disorder, and substance abuse \cite{kalmbach2018impact}. More seriously, they might even commit suicide when their antiwork sentiment becomes unbearably strong, which is even more serious during COVID-19 \cite{awan2022suicide, boxer1995suicide}.

Although few works lay focus on analyzing the workplace, there is research looking into causes and behavior characteristics of negative emotions to one's mental health. Existing work focuses on the psychological reasons for MH problems and treatments for their symptoms like sleep disorder\cite{review_mental_health, kalmbach2018impact}. Some take one's living environment, such as their home and workplace, into account and assess their impact on their mental health \cite{maslach2001job}. De Choudhury et al's work \cite{suicide_reddit}, in particular, shows a systematic way of using mental health content in social media to detect MH problems in society. To overcome the inefficiency of recognizing suicidal ideation in traditional psychological, psychiatric, and demographic examinations, they used Reddit to investigate the insights into the psychological states, health, and well-being of individuals. In addition to linguistic features, they took advantage of the time and inter-subreddit relationship among 14 MH-related subreddits and found more accurate signs of suicide ideation shifts. In this work, we investigate similar techniques in the workplace-related antiwork subreddit, an online forum of work critiques and labor movements that provides a direct point of view on one's opinion to work. This perspective is more beneficial than general work-related forums used in some prior works.

Moreover, the methods of previous research on MH problem can be improved. Most conclusions have to be obtained manually \cite{jiang-etal-2020-detection, low2020natural, Choudhury_De_2014}. This is because they are limited to common natural language processing (NLP) techniques and the insights they proposed heavily rely on human intelligence, which might not be holistic and fair sometimes. As generative AI evolves rapidly, we are able to learn more hidden features and insights that are more beneficial in terms of broadening the solutions to MH problems. 

In this project, our goal is to investigate the behavioral characteristics and causes of antiwork. We first identified those who posted negative sentiments in the antiwork subreddit after they post some neutral content in some other work-related subreddits as ``antiwork'' users. We then ran a robustly optimized BERT pretraining approach (RoBERTa)-based recurrent neural network (RNN) feature extraction model to understand the characteristics of posts that increase the likelihood that someone will become antiwork. We finally summarize antiwork characteristics and leading factors with linguistic inquiry and word count (LIWC) and topic modeling. Our model learns why people hate their work, offering insights into how to reduce the toxicity of workplaces.

Overall, our study has the following contributions:
\begin{itemize}
    \item A RoBERTa-based RNN antiwork feature extraction model with 80\% accuracy. It can also highlight the antiwork parts of a post.
    \item The behavioral characteristics of antiwork users, including linguistic, interpersonal, and interaction features. This could help identify work-related stress, and thus reveal negative workplace in an early stage.
    \item Identification of antiwork characteristics using topic modeling. These findings have a significant benefit to worker rights protection.
\end{itemize}

Our study also has the potential to be generalized to other social problem analyses over social media. Depending on the social problem and the structure of the social media, the specific large language model (LLM) and topic modeling techniques should be tailored to the research target.

\section{Related Works}
Our method builds on prior works in mental health studies, Reddit data analysis, social media analysis with machine learning, and topic modeling on qualitative analysis.
\subsection{Mental Health Problems and Toxic Workplace}
Mental health problems are increasingly being recognized as a critical issue in modern society. Conditions like depression and burnout can lead to sleep disorders, insomnia, and even suicide \cite{review_mental_health, kalmbach2018impact, maslach2001job}. Various factors contribute to mental health problems, and one prevalent factor in today's world is the toxic work environment. Low wages and high pressure at work are two common reasons why people dislike their jobs. Moreover, there are additional detrimental behaviors in the workplace, such as harassment, bullying, and ostracism \cite{limm2011stress, rasool2021toxic, luo2008survey}. Therefore, our goal is to thoroughly examine the impact of toxic workplaces to effectively address the numerous mental health problems associated with them.

\subsection{Antiwork Subreddit}
Antiwork subreddit is an online forum associated with critiques of work and labor movements \cite{antiwork1, antiwork2, antiwork3}. While some workers are able to stand out and protect their rights, many individuals can only express their dissatisfaction online. While they might feel good after explicitly expressing their negative emotions, it is not truly helpful in relieving their mind if the MH problem is not solved \cite{mckenna1999causes}. On the contrary, it might cause emotional contagion, which amplifies their negative emotions and cause mental health problems \cite{kramer2014experimental}. 

Recent years have seen a rapid increase in antiwork subreddit \cite{antiwork_subreddit}. COVID-19 and its economic damages to the world increase unemployment and employers are harsher and harsher to their employees \cite{saladino2020psychological, osofsky2020psychological, chakraborty2020covid}. The antiwork subreddit has become a popular place for people to share their unpleasant working experiences. A large portion of posts are negative, showing signs of depression and hopelessness that are likely to develop into MH problems \cite{hermida2023mental}.

\subsection{Machine Learning Application in Social Media Analysis}
Researchers have used behavioral and linguistic cues from social media data, such as Twitter, Reddit, and Facebook, to study the mental health status of users \cite{chancellor2020methods}. Neural networks and deep learning methods \cite{gkotsis2017characterisation} are increasingly popular in recent years to predict the mental health status of the user behind their posts, while traditional machine learning techniques such as supervised learning, principal component analysis and support vector machines (SVM) are still the popular choice for simpler NLP tasks. In De Choudhury et al's work \cite{suicide_reddit}, the goal is to identify individuals who are at risk for suicide. Thus, De Choudhury et al focused on the most distinguishable characteristics of the two groups. If one starts to post in r/SuicideWatch within a range of time after they post in other MH subreddits, they are identified as suicide prone. In this paper, we focus on insights into the antiwork causes and characteristics. We want to find out what the contributing factors are so that we can make the proper attempts to eliminate the root of antiwork sentiments. Therefore, traditional machine learning techniques might not be enough to find out more in-depth results due to their limited inference ability. More advanced techniques, such as LLM, can be applied to handle more complex feature extraction. Together with topic modeling, its result can be more interpreted and thus we can gain deeper insights.

\subsection{Topic Modeling on Qualitative Analysis}
Latent Dirichlet Allocation (LDA) is a common technique used for topic modeling \cite{nikolenko2017topic, tong2016text, rehurek_lrec}. The idea behind LDA is that each document is a combination of several topics, while topics are represented by words. Thus, by finding the relationship between words and topics, each document can be clustered into topics based on the words in it. This simple and fast topic modeling technique shows great success in qualitative analysis. In our study, we also use LDA to conduct topic modeling and find out the most important aspects that workers feel uncomfortable with.

\section{Data}
Since we are interested in analyzing the behavioral characteristics of antiwork from social network posts, we turn to Reddit, the popular online forum rich in interpersonal discussion and other forms of social interaction for data.

\subsection{Data Collection}
As antiwork ideation is most frequently implied in people's discussions of workplace context, we pick five subreddits that have high relevance, including ``r/antiwork'', ``r/recruiting'', ``r/recruitinghell'', ``r/work'', and ``r/jobs''.
We obtain the raw data via Academic Torrents \footnote{\href{https://academictorrents.com/}{https://academictorrents.com/}}, consisting of a total of 329,830 posts and 400,000 comments (also referred to as ``posts'' for simplicity)\footnote{This work does not distinguish between Reddit posts and comments.}, with a time range of 03/27/2009 - 12/31/2022. The detailed data distribution among different subreddits is shown in Table \ref{tab:dataset-raw} and Fig \ref{fig:data-time}.

\begin{figure}[htb]
    \centering
    \includegraphics[width=.98\linewidth]{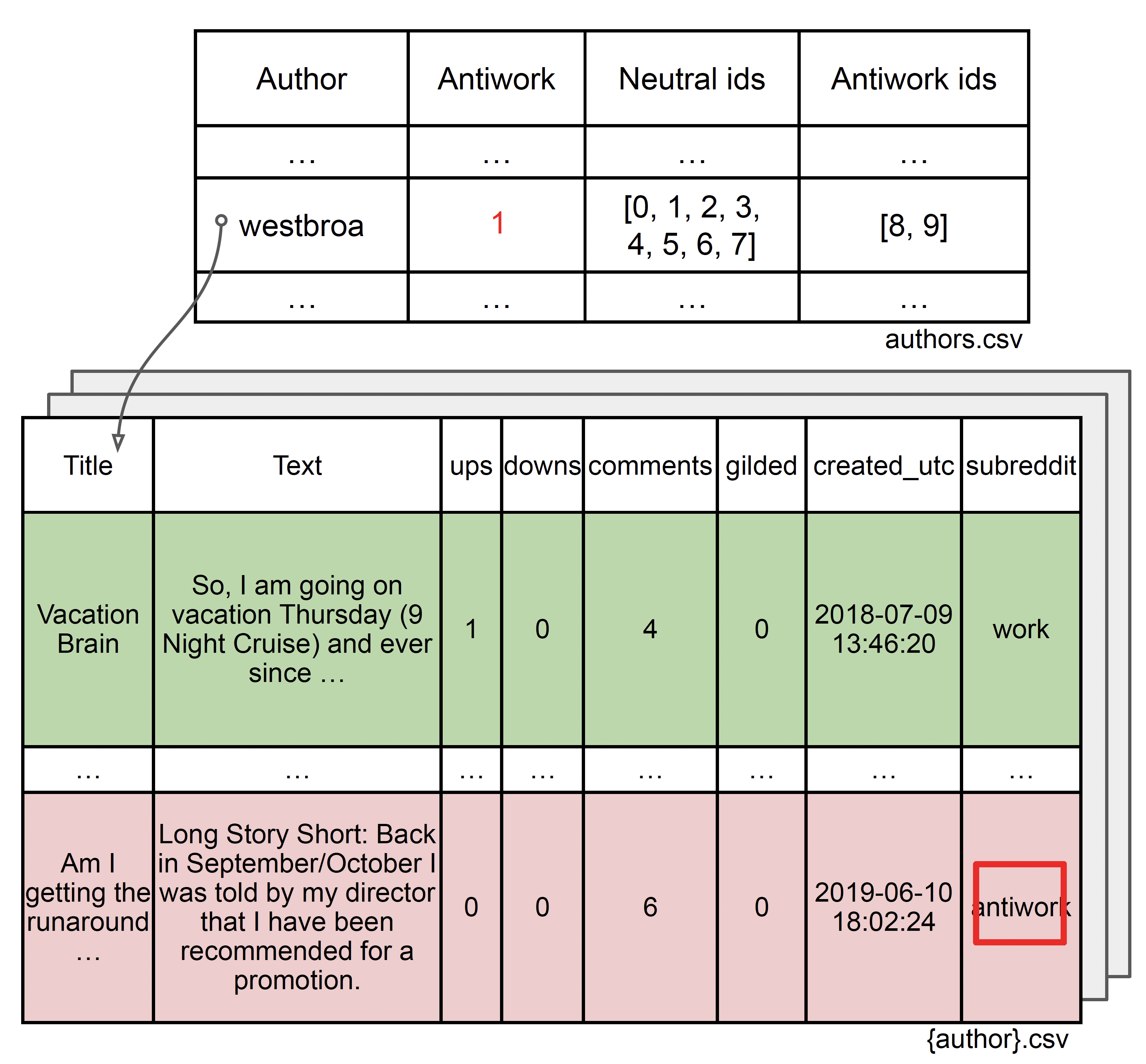}
    \caption{Data structure of Antiwork Reddit Dataset.}
    \label{fig:data}
\end{figure}

The raw datapoints we obtain remain post-level. We hence convert them into more structured, user-level datapoints by grouping the posts that belong to each user and sorting them in chronological order (see Fig \ref{fig:data}). Moreover, we discard datapoints where either ``selftext'' or ``title'' field is missing.

\subsection{Constructing User Classes}
\label{sec:construct_class}
We take a similar approach as \cite{suicide_reddit} to create user groups \{``antiwork'', ``neutral''\} - using subreddits as a proxy. Specifically, we decide that a user has an ``antiwork'' propensity if there exists a r/antiwork post (chronologically) after a post from ``neutral subreddits'' (``r/recruiting'', ``r/recruitinghell'', ``r/work'', and ``r/jobs''); a user tends to be ``neutral'' if the condition above is not met and all the posts come from ``neutral subreddits''. Similar to \cite{suicide_reddit}, this labeling schema serves as an efficient way of large-scale user modeling which would nevertheless induce noises. To mitigate the potential error of a user being labeled ``neutral'' but demonstrating ``antiwork'' emotion outside the ``time periods'' \cite{suicide_reddit}, in our approach we do not confine the time window for different subreddit posts to appear. We instead observe the whole time range and only focus on whether there exists a temporal order. We plot the schema of labeling in Fig \ref{fig:data-labeling}. The post history (time normalized within [0,1]) for users from different classes is also visualized in Fig \ref{fig:data-class-time}.

This ends up with 855 antiwork users and 83872 neutral users. To obtain a balanced class ratio and average post number of each class, we sample 1000 neutral users with Gaussian probabilistic function $\mathcal{N}(\mu_\text{antiwork},\sigma^2_\text{antiwork})(\cdot)$ on the average post number. Before training, we further split the data into training (0.75) and testing (0.25) sets. The final statistics are shown in Table \ref{tab:dataset}.

\subsection{Data Cleaning}
We sanitize the ``title'' and ``selftext'' fields by replacing URLs and numbers with special tokens ``url'' and ``@''), and emojis with their meanings (with the \textit{emoji.demojize} package \cite{Taehoon-Kim-Kevin-Wurster-2023}). For URLs and numbers, we manually examine them and verify that they do not contribute to contexts for understanding antiwork.

\begin{table}[htb]
    \centering
    \resizebox{\linewidth}{!}{
    \huge
    \begin{tabular}{c|c|c}
        &Antiwork & Neutral \\
        &\begin{tabular}{cc}
           \# data  & (mean, std) 
        \end{tabular}&
        \begin{tabular}{cc}
           \# data  & (mean, std) 
        \end{tabular}
        \\\midrule
       \begin{tabular}{c}
        Train (0.75)\\
        Val (0.25)
       \end{tabular} &
       \begin{tabular}{cc}
        \begin{tabular}{c}
           641\\
           214
        \end{tabular} & (7.55, 11.29)
       \end{tabular} &
       \begin{tabular}{cc}
        \begin{tabular}{c}
           750\\
           250
        \end{tabular} & (6.53, 4.65)
       \end{tabular}
    \end{tabular}
    }
    \caption{Statistics of Antiwork Reddit Dataset.}
    \label{tab:dataset}
\end{table}
\section{Methods}
In order to understand the underlying causes of antiwork emotions, we take two different approaches. On the one hand, we experiment with different models for antiwork propensity prediction, whose features can be interpreted and help to trace back antiwork-related factors. On the other hand, we directly analyze the data, seeking linguistic patterns that may embody antiwork emotions.

\subsection{Feature Extraction Model}
\label{sec:model}

The training process for our model is set up as given the posts of a user, a label of this user is predicted. The models we experiment with include:

\paragraph*{SVM: TF-IDF}
The Support Vector Machine (SVM) is a linear model frequently used for simple classification tasks. In this case, we take the term frequency-inverse document frequency (TF-IDF) over all the posts a user has as the feature, which could give us clues on how antiwork is correlated with word occurrence. 

\paragraph*{SVM: Linguistic \& Social Engagement Features}
Besides word-level features, we also try features that tend to be more abstractive, namely, linguistic patterns (\#first/second/third-person singular/plural, \#noun, \#verb, \#adj, \#adv, \#cconj, \#num, \#punct, and \#pron) and social engagement (\#score, \#ups, \#downs, \#comments, \#gilded, and \#pinned), for the SVM backbone. The linguistic features are obtained with spaCy. The social engagement features are directly imported from metadata. 

\paragraph*{BERT (base)}
BERT is a transformer-based encoder that can transform a long text into a content-based feature vector of length 768. Here we concatenate all the posts of a user before passing them to BERT for embedding, which is then passed to a linear prediction head.

\begin{figure}
    \centering
    \includegraphics[width=.45\textwidth]{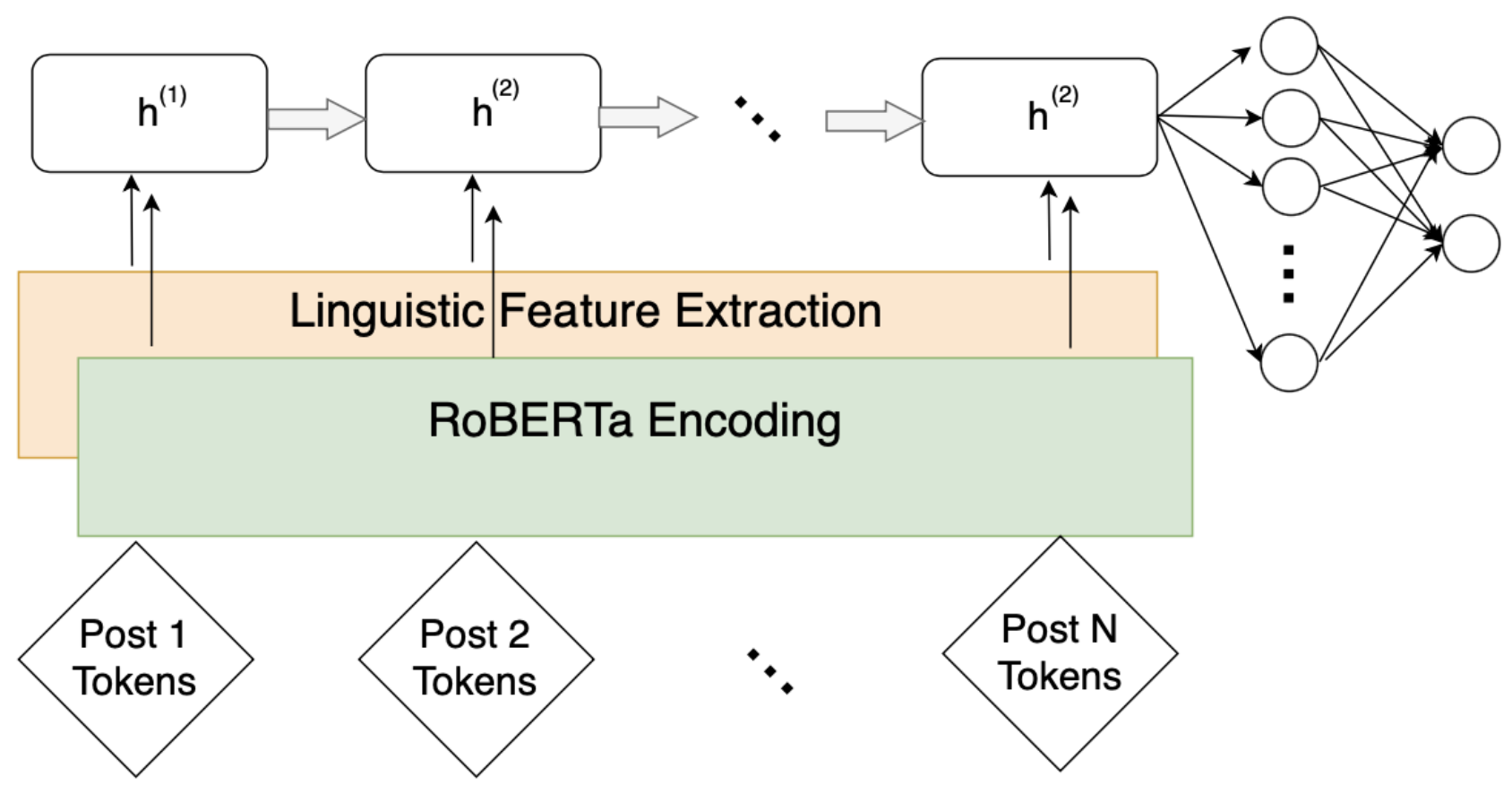}
    \caption{Architecture of our model}
    \label{fig_model}
\end{figure}

\paragraph*{RoBERTa + RNN}
RoBERTa is a robustly optimized BERT model that also encodes the text. For each post, we first obtain its RoBERTa embedding and concatenate it with its linguistic feature vector. We then use an RNN to further incorporate all the post vectors belonging to a user in series, aiming to capture the dependency between posts. The hidden state of the last node in the RNN is sent to a fully connected layer for antiwork scores. The architecture is shown in Fig \ref{fig_model}.

After training, we utilize the models to infer the correlation between features and antiwork.

For SVM-based methods, we investigate the weights corresponding to each feature, indicating antiwork score contribution. This makes sense as the values are normalized among the features.

\begin{figure}
    \centering
    \includegraphics[width=.45\textwidth]{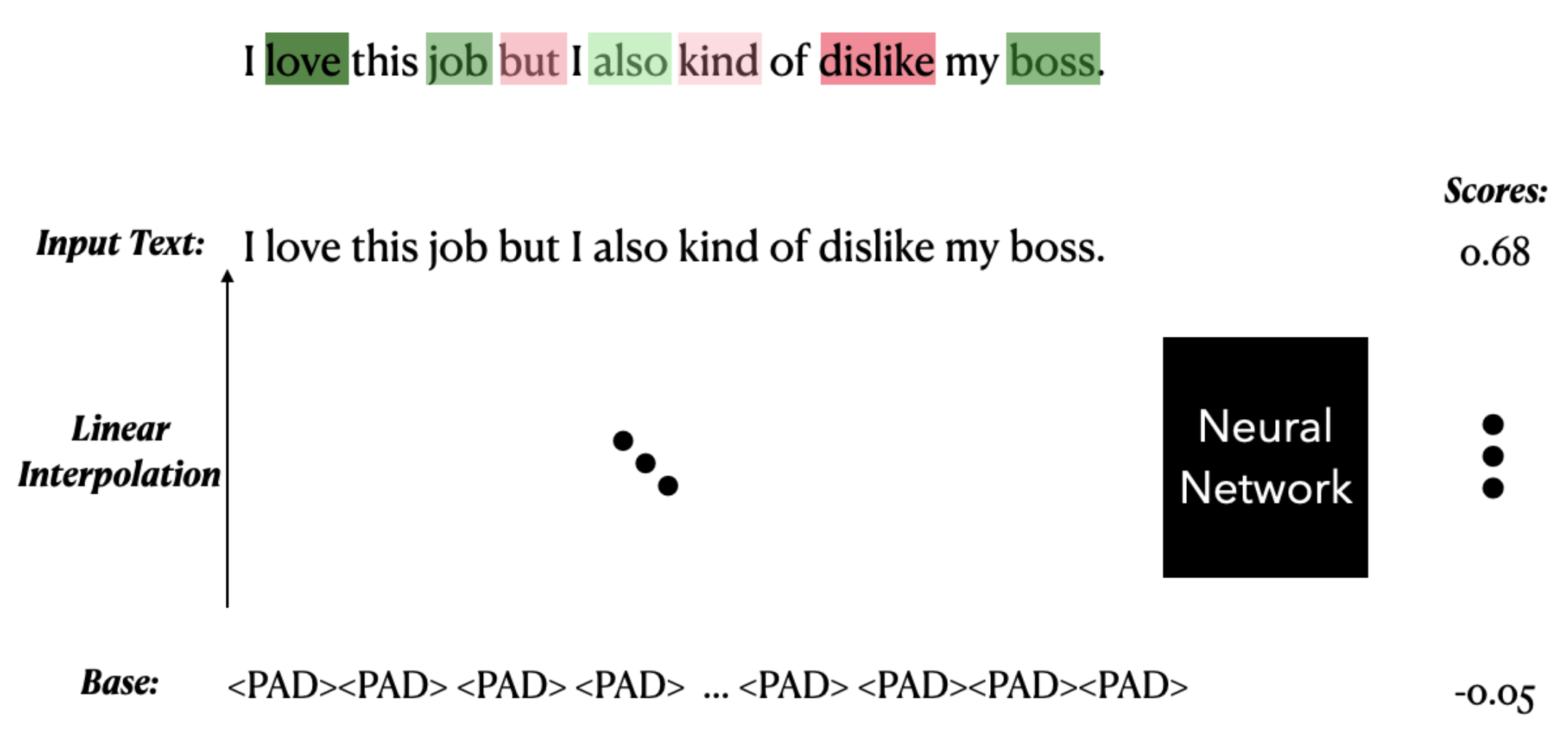}
    \caption{Overview of the application of integrated gradient. Red for words that positively contribute to antiwork prediction and green for words that negatively contribute to antiwork prediction.}
    \label{fig_IG}
\end{figure}

For RoBERTa-RNN method, we perform word-level attribution (see Fig \ref{fig_IG}). With the model, each word is given a score, and higher values indicate a higher relationship with the source of antiwork. We visualize the importance of each input word that contributes to the model's prediction by a technique called integrated gradient \cite{sundararajan2017axiomatic}. Starting from a baseline, which is an empty sentence, a linear interpolation between the baseline and the input text is generated. Then by measuring the relationship between changes to a feature and changes in the model’s predictions, the importance of each feature can be determined.

For the computation of integrated gradient, we used a library based on Pytorch, called Captum \cite{kokhlikyan2020captum}.

\subsection{Feature Analysis}
We further conduct separate analyses on the posts of different user classes with statistical methods.

\paragraph*{Antiwork characteristics}
We use LIWC \cite{boyd2022development} to extract the word count and linguistic features (number of occurrences) of different categories for Reddit posts from both ``antiwork'' and ``neutral'' users. For each feature, we run a Wilcoxon signed-rank test to evaluate if the feature shows a statistically significant difference between the two groups of users. The results are presented in Table \ref{tab:LIWC}. 

\paragraph*{Latent Dirichlet Analysis}
To find out the leading factors for antiwork, we analyze the dataset with both our model and a topic modeling model \cite{rehurek_lrec}, and reveal the topic in antiwork posts. We use the \texttt{gensim} package \cite{rehurek_lrec} to group posts based on their topics, which are mostly dissatisfactions people complain about. The package uses LDA to learn the optimal clustering of topics in the text and identifies keywords in each topic. We also use our model discussed in Sec \ref{sec:model} to find out some typical negative antiwork posts with high weights in the sentences. We use the highlighted words to check if the topic modeling technique returns reasonable groupings.

\section{Result}
\subsection{Feature Extraction Model}
\paragraph*{Predictive model performance} The results are shown in Table \ref{tab:perform}. When considering only the post content, both TF-IDF and Bert embeddings achieve about 0.65 accuracies. While the model using Bert embedding has an F-1 score of 66\%, TF-IDF only achieved 56\%. This indicates that linguistic features other than word choices are also distinguishable between antiwork users and others. In fact, the SVM models using linguistic features and social engagement features alone give 68\% accuracy and 64\% F-1 score. 

Our best model, using RoBERTa combined with an RNN, incorporates both LLM embedding in the text and other linguistic and social engagement features, achieving $80\%$ accuracy and $0.79$ F-1 score. We use this model for the rest of our findings.

\begin{figure} [ht]
    \centering
    \includegraphics[width=.45\textwidth]{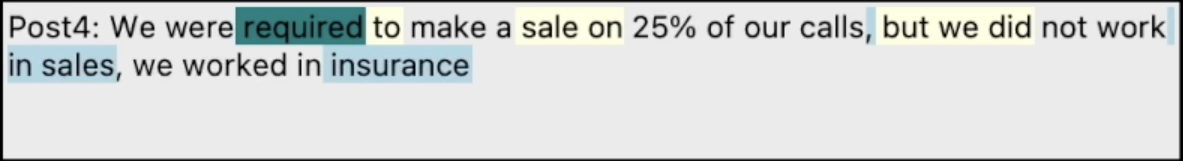}
    \caption{Visualization of word attributions}
    \label{fig:post4}
\end{figure}

\begin{table*}[htbp]
\centering
\begin{tabular}{c|cccc}
Model                                      & Accuracy & Precision & Recall & F1-score \\ \hline
Random (baseline)                          & 0.54     & 0.46      & 0.50   & 0.48     \\
SVM: TF-IDF                                & 0.64     & 0.65      & 0.50   & 0.56     \\
SVM: Linguistic Social Engagement Features & 0.68     & 0.67      & 0.62   & 0.64     \\
BERT (base)                                       & 0.68     & 0.66      & 0.67   & 0.66     \\
\textbf{RoBERTa + RNN}                              &
\textbf{0.80}     & \textbf{0.83}      & \textbf{0.75}   & \textbf{0.79}
\end{tabular}
\caption{Preliminary experiment results for antiwork propensity prediction.}
\label{tab:perform}
\end{table*}
\paragraph*{Visualization Interface}
A visualization interface is created to visualize word attributions and thus help post-analysis. Based on the attributions returned by the model, the interface highlights words with different background colors. Specifically, light yellow, yellow, and red represent positive attributions from low to high, while light blue, teal and blue represent negative attributions from low to high (see Fig  \ref{fig:post4}). 

\begin{figure} [ht]
    \centering
    \includegraphics[width=.45\textwidth]{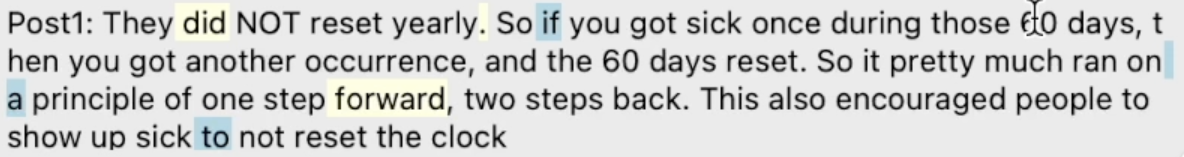}
    \label{fig:post1}
\end{figure}

\subsection{Linguistic, Interpersonal, \& Interaction Analysis}

\subsubsection{Antiwork characteristics}
We present some distinct characteristics of users labeled ``Antiwork'' based on LIWC results. Overall, they are less confident about themselves and careless about their life.

\paragraph*{Lack of Self-affirmation}
According to the result of Wilcoxon signed-rank test on LIWC features, there exist statistically significant differences between the two user groups in the scores of tones and emotions - ``antiwork'' users tend to sound more negative and suspicious in their Reddit posts. On the one hand, they tend to be more effusive of emotions in their tone ($z_\text{tone}=-13.181$), most of which are negative ($z_\text{tone\_pos}=-6.113$, $z_\text{emo\_pos}=-4.350$). On the other hand, they tend to use more question marks in the text ($z_\text{QMark}=-14.065$).

\paragraph*{Carelessness about life}
We also find statistically significant differences in ``work-life'' topic word occurrence. While ``antiwork'' users talk more about ``work''-related topics ($z_\text{work}=-20.804$), they care less about daily-life topics such as physical wellness ($z_\text{physical}=-7.693$), mental wellness ($z_\text{mental}=-1.385$) and food ($z_\text{mental}=-2.843$), disclose less their desires ($z_\text{want}=-3.998$), and focus less on the present ($z_\text{focuspresent}=-7.384$). These factors demonstrate that antiwork sentiment could badly distract people from enjoying their current lives. 
\begin{table*}[t]
\centering
\begin{adjustbox}{width=.8\textwidth}
\Large
\begin{tabular}{@{}ccccccc@{}}
\toprule
\multicolumn{2}{c}{Category}                            & Antiwork & Neutral & z       & p    \\\midrule
Summary Variables                        & \textbf{Tone}         & \textbf{38.094}   & \textbf{32.419}  & \textbf{-13.181} & \textbf{***}  \\\cmidrule(l){1-2} 
Linguistic                               & ipron        & 4.184    & 5.108   & -4.811  & **   \\\cmidrule(l){1-2} 
\multirow{7}{*}{Psychological Processes} & Drives       & 4.951    & 5.926   & -5.179  & **   \\
                                         & discrep      & 1.919    & 1.840   & -6.140  & ***  \\
                                         & tentat       & 2.502    & 2.293   & -9.017  & ***  \\
                                         & Affect       & 4.527    & 6.341   & -8.228  & ***  \\
                                         & \textbf{tone\_pos}    & \textbf{2.824}    & \textbf{2.948}   & \textbf{-6.113}  & \textbf{***}  \\
                                         & \textbf{emo\_pos}     & \textbf{0.588}    & \textbf{0.655}   & \textbf{-4.350}  & \textbf{***}  \\
                                         & swear        & 0.298    & 0.834   & -7.828  & ***  \\\cmidrule(l){1-2} 
\multirow{11}{*}{Expanded Dictionary}    & money        & 1.361    & 1.811   & -4.001  & *    \\
                                         & \textbf{work}         & \textbf{11.030} & \textbf{7.574} & \textbf{-20.804} & \textbf{***} \\
                                         & \textbf{Physical}     & \textbf{0.949}  & \textbf{1.679} & \textbf{-7.693}  & \textbf{***} \\
                                         & \textbf{illness}      & \textbf{0.096}  & \textbf{0.276} & \textbf{-2.848}  & \textbf{***} \\
                                         & \textbf{mental}       & \textbf{0.047}  & \textbf{0.139} & \textbf{-1.385}  & \textbf{*}   \\
                                         & \textbf{food}         & \textbf{0.188}  & \textbf{0.299} & \textbf{-2.843}  & \textbf{***} \\
                                         & \textbf{want}         & \textbf{0.388}  & \textbf{0.420} & \textbf{-3.998}  & \textbf{***} \\
                                         & \textbf{allure}       & \textbf{6.749}  & \textbf{7.930} & \textbf{-6.854}  & \textbf{***} \\
                                         & \textbf{focuspresent} & \textbf{4.652}  & \textbf{5.815} & \textbf{-7.384}  & \textbf{***} \\
                                         & Comma                 & 2.481           & 2.359          & -4.635           & **           \\
                                         & \textbf{QMark}        & \textbf{1.917}  & \textbf{1.844} & \textbf{-14.065} & \textbf{***} \\ \bottomrule
                                         *: 0.05/N, **: 0.001/N, ***: 0.001/N
\end{tabular}
\end{adjustbox}
\caption{Difference of LIWC results between antiwork and neutral posts.}
\label{tab:LIWC}
\end{table*}

\subsubsection{Leading Factors for Antiwork}
\begin{figure}[htbp]
    \centering
    \includegraphics[width=0.5\textwidth]{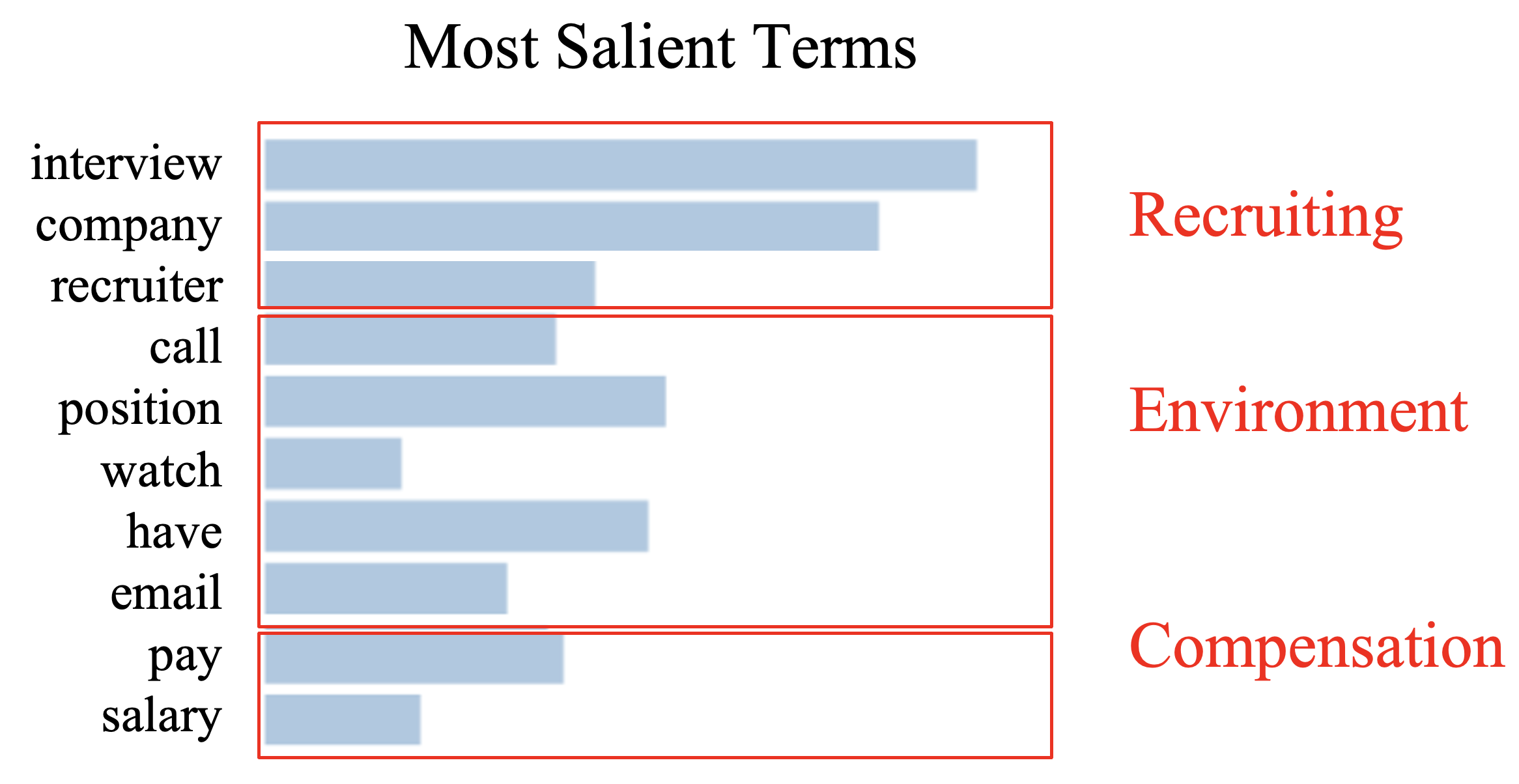}
    \caption{Top-10 most salient terms from Topic Modeling. Each row shows the saliency of a certain word to the dataset. The longer the bar in the row is, the more salient the word is.}
    \label{fig:topic_modeling}
\end{figure}

Using the LDA technique described in Section 4.2, we determine the top salient terms from topic modeling which can be seen in Fig \ref{fig:topic_modeling}. Using these topics, we find three broad categories of causes - recruiting, work environments, and compensation. We perform a qualitative analysis of posts grouped under these topics to better understand the key factors that cause antiwork sentiments. 

\paragraph*{Harsh Working Environments}
Antiwork sentiments arise when workers feel unvalued, untrustworthy, and not respected or welcomed. One user stated \textit{``I have fewer motivations for work because I feel like I don’t have influence. Kinda feel like my opinion/minds are not as important as my ability to just get things done.''} It is especially the case when they are iced out by co-workers or are suspected of doing evil. Another user described the experience of being accused of accepting counterfeit money and being asked to pay the amount out of pocket. They said \textit{``No one at my job likes me so I'm wondering if I should just pay the money.''}

Some users were burdened by stressful environments that put a lot of pressure on employees. Some of them were required never to make mistakes, making them feel over-stressful. One user shared that they would be punished if anyone in the team made a tiny mistake in a huge task containing plenty of tasks to do - \textit{``If anyone did things wrong, my head will be on the chopping block. But there are countless tiny details to remember and you have to be damn near perfect to have every move go smoothly.''}

\paragraph*{Frustrating Recruiting Experiences}
A surprising cause of the antiwork sentiment is the difficulty that users experienced while trying to find jobs. Job hunting can cause anxiety, especially when peers have already received offers. One user posted, \textit{``Depressed over the internship hunt, especially when everyone in my school has offers already. I’ve felt like a loser all the time in my cursed, meaningless life. I would kill myself in my dream if I could''}. According to the 2017 Talent Acquisition Benchmarking Report from SHRM, the average length of the hiring process is 36 days \cite{SHRM}. Many workers are exhausted from the lengthy recruiting process, which only become more frustrating when the companies take the offer back or impose additional requirements. One user recorded the process of them being accepted and then rejected suddenly on Reddit, \textit{``After three interviews, the company said they would offer me a job a month ago, but rejected me yesterday. Now the recruiter has posted the same job again. I'm tired of recruiters playing games with me.''} Making the process of being accepted into a working environment more challenging than it needs to be leads to frustration and discourages individuals, thereby causing the antiwork sentiment to manifest. 

\paragraph*{Unfair Compensation}
Low wages and long work hours are two commonly mentioned problems, with unfair remuneration being a major source of dissatisfaction. Some users mentioned that their wages had remained the same since the 90s with few paid leaves, and even sickness was not an exception. To paraphrase one user -  \textit{``Many of the packing places were only offering \$$15$-\$$16$ an hour. A lot of places use the "up to" crap.''} Another talked about how their workplace had unfairly high expectations of them without offering sufficient compensation, saying, \textit{``Our coach wants us to work like team leads, but get paid less than team leads. Why do I have to work like a team lead, but still be paid as such?''}

Restricted leave pay is another aspect of unsatisfying treatment. As one user mentioned: \textit{``the occurrences fall off, ONE per 60 days. Meaning you pretty much had a two-month probation of not being sick, your car starting every day, not being late, etc.''} Workers with low wages are more likely to have restricted leave pay. According to the data on employee benefits released by the Bureau of Labor Statistics, more than 60\% of the lowest-paid employees in the United States are unable to receive paid sick leave to take care of themselves or their family members \cite{laborbureau}. 

\section{Discussion}
Lastly, we provide discussions on the implications our study may have or be subjected to, in terms of ethical considerations and future work.

\subsection{Ethical Implications} 

While our objective is to uncover underlying causes of workplace dissatisfaction and employee unhappiness, we realize that our solution does have ethical implications. A classifier that can detect if a person is a member of the antiwork movement has the potential to be misused to stop strikes, fight against unions or falsely accuse workers of being against the company. To prevent this form of misuse, we will not be making the actual model or complete dataset available. We only share the anonymized antiwork labeled data to prevent training a classifier, and we do not share the trained classifier, but rather, just the word attribution utility. By doing so, we wish to prevent the misuse of this technology, since it can no longer classify individuals, but can be used to identify potential problems in the workplace. We also understand that although the posts used in our dataset are public, users may not be willing to have their personal information tracked and analyzed. To mitigate the problem, we have removed all personally identifiable information in the dataset and de-identified and paraphrased all the texts before presenting them. We encourage any further research involving this form of data to do the same. 


\subsection{LLM on Social Media Analysis}
In our work, we send RoBERTa-generated encodings to RNN to learn the difference between the posts sent by antiwork users and neutral users. The model achieves 80\% overall accuracy, and performs well on both classes. Our study shows the great potential of LLM in social media analysis. As an emergent technology, LLM takes advantage of its large pre-train dataset and deep neural network, providing a fairly comprehensive understanding of natural language. Concatenated with classification layers, 
LLM would be extremely powerful in terms of feature extractions and predictions. This could be an effective model to process articles like social media posts on Reddit, and provide insights into them. We suggest future works to investigate the usage of LLM on more social phenomena analysis tasks like this. 

\section{Conclusion}
In this work, we present and describe the creation of a dataset using Reddit data aimed at uncovering the underlying causes of workplace dissatisfaction and antiwork sentiments. Using a RoBERTa feature extractor and an RNN model, we can detect users with antiwork sentiments with 80\% accuracy. More importantly, this model suggested that a lack of self-affirmation and carelessness about life indicate workplace dissatisfaction. The model also suggests that harsh working environments, frustrating recruiting experiences and unfair compensation to be the leading causes of negative emotions. After extensive quantitative and qualitative analysis, we find evidence that suggests that individuals dissatisfied with their working environments lack self-confidence and motivation. Our analysis also suggests that the leading causes of antiwork sentiments among posters on the antiwork forum include frustrating recruiting experiences, unfair compensation, and harsh working environments where employees feel friendless or powerless. 

To prevent this work from being misused, we make available only the data corresponding to users classified as antiwork, and our feature extraction model.  We hope that through this research we can help businesses create better practices to improve worker satisfaction and happiness by identifying the root cause of employee unhappiness.

\section*{Limitations}
Since our dataset only comprises text from Reddit posts, we acknowledge that there may be biases due to the linguistic structure of these posts and that our model may not generalize well to data drawn from other forums, social media platforms, spoken language, or formal communication.  Although we hypothesize that similar results can be concluded from other social media posts, future research into other sources of worker sentiment, particularly from formal settings like exit interviews is needed. We also acknowledge that our insights into the recruitment process causing antiwork sentiments may be biased by our choice of subreddits. 

Also, as discussed in Sec \ref{sec:construct_class}, because of the way we identify our user groups and the corresponding labels, we are only able to approximately describe users' antiwork propensity, which would induce noises within each group.  Workplace dissatisfaction is a subtle, often mixed feeling, and incorporates multiple emotions such as anger, frustration, fatigue, and anxiety. Our approach is unable to distinguish these subtle emotions or break sentiments down into these groups. A more accurately labeled dataset using self-reporting would overcome this issue.

\bibliography{anthology,custom}
\bibliographystyle{acl_natbib}

\onecolumn
\newpage
\appendix

\section{Data Time Distribution}
\label{sec:appendix-data}

\begin{table}[htb]
    \centering
    \begin{tabular}{c|c|c}
      Subreddit & \#Posts & \#Comments \\ \hline
      recruitinghell & 54086 & 100000\\
    recruiting & 17370 & 100000\\
    work & 58374 & 100000\\
    jobs & 100000 & 100000\\
    antiwork & 100000 & 0
    \end{tabular}
    \caption{Raw data distribution (posts and comments) for different Subreddits from Academic Torrents.}
    \label{tab:dataset-raw}
\end{table}

\begin{figure}[ht]
    \centering
    \includegraphics[width=\linewidth]{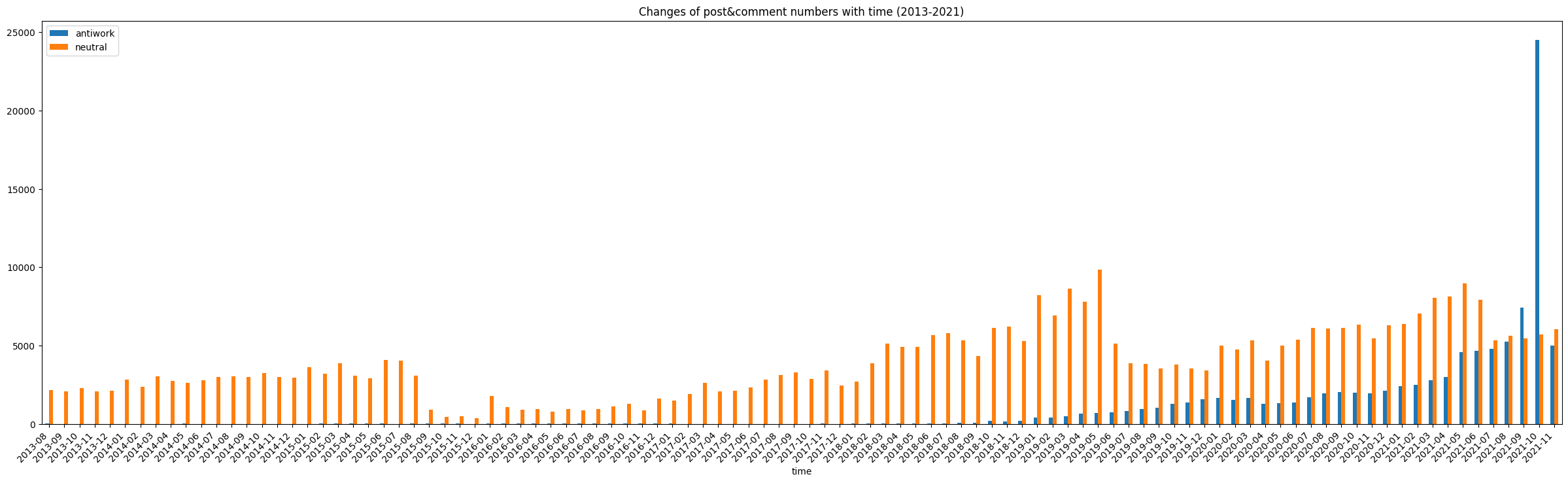}
    \caption{Time distribution of raw data from Academic Torrents.}
    \label{fig:data-time}
\end{figure}

\begin{figure}[ht]
    \centering
    \includegraphics[width=\linewidth]{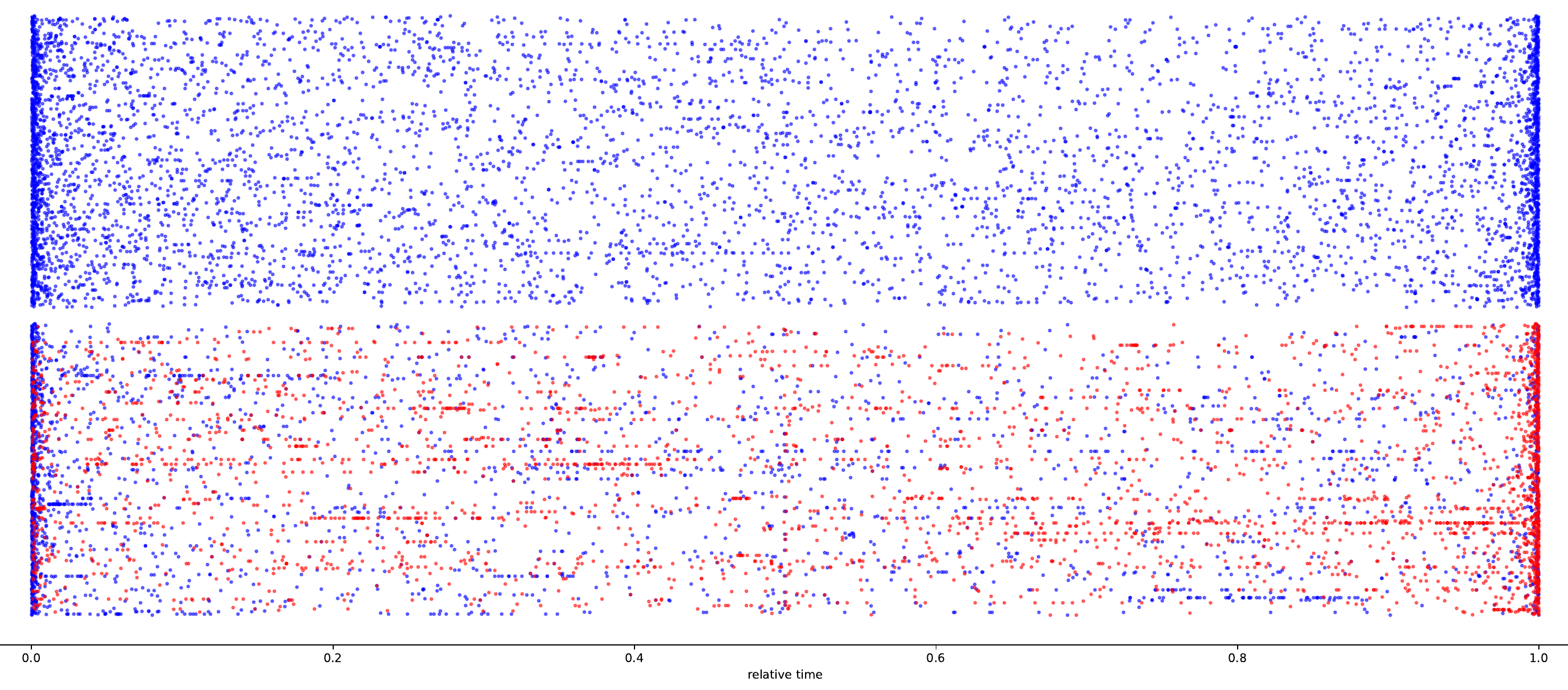}
    \caption{Distribution of post time for different user groups.}
    \label{fig:data-class-time}
\end{figure}

\begin{figure}[ht]
    \centering
    \includegraphics[width=.98\linewidth]{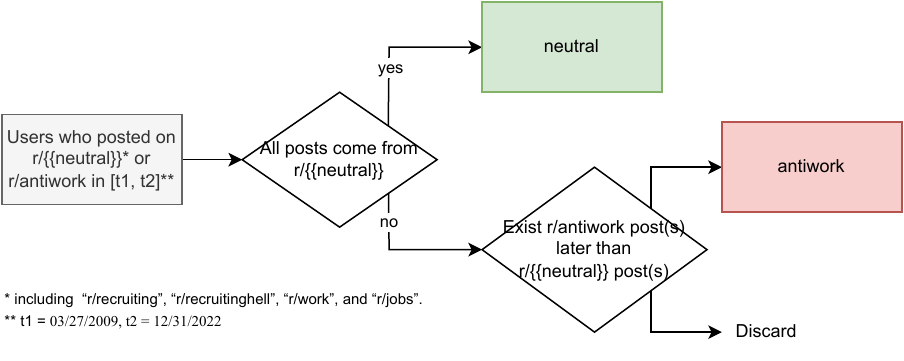}
    \caption{Labeling schema.}
    \label{fig:data-labeling}
\end{figure}

\newpage
\section{LIWC Results (Full)}
\label{sec:appendix-stats}
\begin{table*}[ht]
\centering

\minipage{0.46\textwidth}
\resizebox{\linewidth}{!}{
\begin{tabular}{@{}cccccc@{}}
\toprule
Category     & Antiwork & Neutral & z       & p-value & p   \\\midrule
Segment      & 1.000    & 1.000   & 0.000   & 1.000   &     \\
WC           & 96.316   & 79.982  & -10.398 & 0.000   & *** \\
Analytic     & 54.348   & 48.326  & -10.816 & 0.000   & *** \\
Clout        & 37.295   & 38.838  & -1.237  & 0.212   &     \\
Authentic    & 57.488   & 54.587  & -3.240  & 0.001   &     \\
Tone         & 38.094   & 32.419  & -13.181 & 0.000   & *** \\
WPS          & 13.098   & 12.551  & -6.962  & 0.000   & *** \\
BigWords     & 23.325   & 20.040  & -14.581 & 0.000   & *** \\
Dic          & 86.097   & 88.752  & -11.159 & 0.000   & *** \\
Linguistic   & 61.820   & 66.665  & -11.650 & 0.000   & *** \\
function     & 47.516   & 50.798  & -7.202  & 0.000   & *** \\
pronoun      & 12.326   & 13.976  & -7.356  & 0.000   & *** \\
ppron        & 8.142    & 8.869   & -2.588  & 0.009   &     \\
i            & 4.709    & 4.680   & -3.567  & 0.000   &     \\
we           & 0.360    & 0.714   & -3.085  & 0.000   &     \\
you          & 1.737    & 1.861   & -2.391  & 0.006   &     \\
shehe        & 0.431    & 0.384   & -3.388  & 0.000   & *** \\
they         & 0.759    & 1.037   & -1.293  & 0.121   &     \\
ipron        & 4.184    & 5.108   & -4.811  & 0.000   & *   \\
det          & 11.630   & 13.261  & -8.687  & 0.000   & *** \\
article      & 5.616    & 5.887   & -0.393  & 0.690   &     \\
number       & 0.547    & 0.668   & -2.877  & 0.000   &     \\
prep         & 12.643   & 11.932  & -5.353  & 0.000   & **  \\
auxverb      & 7.654    & 8.718   & -6.040  & 0.000   & *** \\
adverb       & 4.924    & 5.828   & -3.594  & 0.000   &     \\
conj         & 4.817    & 4.817   & -1.720  & 0.078   &     \\
negate       & 1.421    & 1.989   & -3.513  & 0.000   &     \\
verb         & 15.422   & 17.120  & -8.394  & 0.000   & *** \\
adj          & 5.954    & 6.474   & -0.235  & 0.812   &     \\
quantity     & 3.749    & 4.648   & -3.764  & 0.000   &     \\
Drives       & 4.951    & 5.926   & -5.179  & 0.000   & **  \\
affiliation  & 1.010    & 1.453   & -0.499  & 0.551   &     \\
achieve      & 2.323    & 2.428   & -0.560  & 0.549   &     \\
power        & 1.715    & 2.110   & -1.039  & 0.248   &     \\
Cognition    & 12.346   & 13.050  & -0.476  & 0.632   &     \\
allnone      & 0.940    & 1.501   & -1.189  & 0.173   &     \\
cogproc      & 11.315   & 11.464  & -2.636  & 0.008   &     \\
insight      & 3.070    & 2.892   & -8.946  & 0.000   & *** \\
cause        & 1.910    & 1.941   & -3.938  & 0.000   &     \\
discrep      & 1.919    & 1.840   & -6.140  & 0.000   & *** \\
tentat       & 2.502    & 2.293   & -9.017  & 0.000   & *** \\
certitude    & 0.547    & 0.776   & -0.957  & 0.222   &     \\
differ       & 3.010    & 3.142   & -2.553  & 0.008   &     \\
memory       & 0.046    & 0.082   & -0.169  & 0.579   &     \\
Affect       & 4.527    & 6.341   & -8.228  & 0.000   & *** \\
tone\_pos    & 2.824    & 2.948   & -6.113  & 0.000   & *** \\
tone\_neg    & 1.386    & 2.555   & -9.001  & 0.000   & *** \\
emotion      & 1.264    & 1.862   & -0.446  & 0.614   &     \\
emo\_pos     & 0.588    & 0.655   & -4.350  & 0.000   & *** \\
emo\_neg     & 0.608    & 1.044   & -2.828  & 0.000   &     \\
emo\_anx     & 0.121    & 0.175   & -0.821  & 0.086   &     \\
emo\_anger   & 0.129    & 0.229   & -1.901  & 0.000   &     \\
emo\_sad     & 0.094    & 0.118   & -0.282  & 0.460   &     \\
swear        & 0.298    & 0.834   & -7.828  & 0.000   & *** \\
Social       & 12.770   & 12.232  & -5.899  & 0.000   & *** \\
socbehav     & 6.234    & 5.055   & -14.818 & 0.000   & *** \\
prosocial    & 0.810    & 0.656   & -10.026 & 0.000   & *** \\
polite       & 0.339    & 0.284   & -9.380  & 0.000   & *** \\
conflict     & 0.261    & 0.387   & -2.003  & 0.001   &     
\end{tabular}
}
\endminipage
\hfill
\minipage{0.46\textwidth}
\resizebox{\linewidth}{!}{
\begin{tabular}{@{}cccccc@{}}
\toprule
Category     & Antiwork & Neutral & z       & p-value & p   \\\midrule
moral        & 0.275    & 0.418   & -1.929  & 0.001   &     \\
comm         & 2.973    & 2.407   & -12.014 & 0.000   & *** \\
socrefs      & 6.192    & 6.866   & -1.714  & 0.082   &     \\
family       & 0.107    & 0.148   & -0.432  & 0.281   &     \\
friend       & 0.040    & 0.100   & -0.108  & 0.733   &     \\
female       & 0.310    & 0.262   & -2.740  & 0.000   & **  \\
male         & 0.417    & 0.529   & -2.431  & 0.000   &     \\
Culture      & 1.752    & 1.203   & -10.526 & 0.000   & *** \\
politic      & 0.391    & 0.441   & -1.493  & 0.005   &     \\
ethnicity    & 0.115    & 0.112   & -0.027  & 0.916   &     \\
tech         & 1.248    & 0.653   & -11.093 & 0.000   & *** \\
Lifestyle    & 12.240   & 9.753   & -12.927 & 0.000   & *** \\
leisure      & 0.227    & 0.407   & -3.155  & 0.000   & *** \\
home         & 0.125    & 0.215   & -2.732  & 0.000   & *** \\
work         & 11.030   & 7.574   & -20.804 & 0.000   & *** \\
money        & 1.361    & 1.811   & -4.001  & 0.000   & *   \\
relig        & 0.064    & 0.225   & -2.366  & 0.000   & *** \\
Physical     & 0.949    & 1.679   & -7.693  & 0.000   & *** \\
health       & 0.401    & 0.687   & -4.556  & 0.000   & *** \\
illness      & 0.096    & 0.276   & -2.848  & 0.000   & *** \\
wellness     & 0.025    & 0.038   & -0.605  & 0.001   &     \\
mental       & 0.047    & 0.139   & -1.385  & 0.000   &     \\
substances   & 0.004    & 0.019   & -0.190  & 0.165   &     \\
sexual       & 0.041    & 0.068   & -0.697  & 0.000   &     \\
food         & 0.188    & 0.299   & -2.843  & 0.000   & *** \\
death        & 0.091    & 0.186   & -1.819  & 0.000   & *   \\
need         & 0.661    & 0.679   & -0.369  & 0.621   &     \\
want         & 0.388    & 0.420   & -3.998  & 0.000   & *** \\
acquire      & 1.047    & 1.037   & -4.903  & 0.000   & *** \\
lack         & 0.194    & 0.273   & -0.699  & 0.183   &     \\
fulfill      & 0.146    & 0.202   & -0.365  & 0.522   &     \\
fatigue      & 0.079    & 0.193   & -2.354  & 0.000   & *** \\
reward       & 0.349    & 0.299   & -2.574  & 0.000   &     \\
risk         & 0.286    & 0.272   & -0.720  & 0.206   &     \\
curiosity    & 0.644    & 0.469   & -8.033  & 0.000   & *** \\
allure       & 6.749    & 7.930   & -6.854  & 0.000   & *** \\
Perception   & 8.310    & 7.567   & -8.181  & 0.000   & *** \\
attention    & 0.787    & 0.538   & -7.925  & 0.000   & *** \\
motion       & 1.065    & 1.183   & -2.061  & 0.019   &     \\
space        & 5.648    & 4.971   & -8.649  & 0.000   & *** \\
visual       & 0.753    & 0.728   & -5.719  & 0.000   & *** \\
auditory     & 0.167    & 0.183   & -1.618  & 0.002   &     \\
feeling      & 0.319    & 0.413   & -2.174  & 0.001   &     \\
time         & 4.420    & 4.746   & -0.794  & 0.413   &     \\
focuspast    & 3.018    & 3.079   & -3.518  & 0.000   &     \\
focuspresent & 4.652    & 5.815   & -7.384  & 0.000   & *** \\
focusfuture  & 1.063    & 1.226   & -2.751  & 0.002   &     \\
Conversation & 0.734    & 1.308   & -0.101  & 0.892   &     \\
netspeak     & 0.428    & 0.929   & -0.012  & 0.985   &     \\
assent       & 0.235    & 0.372   & -1.075  & 0.043   &     \\
nonflu       & 0.121    & 0.132   & -0.019  & 0.946   &     \\
filler       & 0.028    & 0.055   & -0.198  & 0.278   &     \\
AllPunc      & 21.332   & 22.882  & -1.647  & 0.099   &     \\
Period       & 5.088    & 6.417   & -1.755  & 0.071   &     \\
Comma        & 2.481    & 2.359   & -4.635  & 0.000   & **  \\
QMark        & 1.917    & 1.844   & -14.065 & 0.000   & *** \\
Exclam       & 0.525    & 1.211   & -0.251  & 0.680   &     \\
Apostro      & 2.216    & 3.023   & -3.797  & 0.000   &     \\
OtherP       & 9.106    & 8.028   & -8.482  & 0.000   & *** \\
Emoji        & 0.000    & 0.000   & 0.000   & 1.000   &    \\\bottomrule\\
\end{tabular}
}
\endminipage
\caption{Full LIWC results for antiwork and neutral submission comparisons (*: 0.05/N, **: 0.001/N, ***: 0.001/N).}
\label{tab:scores}
\end{table*}

\end{document}